%% file: main.tex
\PassOptionsToPackage{table,dvipsnames,svgnames,x11names}{xcolor}
\documentclass[10pt,twocolumn,letterpaper]{article}

\usepackage{cvpr}              
\usepackage[accsupp]{axessibility}
\input{preamble}
\definecolor{cvprblue}{rgb}{0.21,0.49,0.74}
\usepackage[pagebackref,breaklinks,colorlinks,allcolors=cvprblue]{hyperref}

\hypersetup{
    colorlinks=true,
    linkcolor=blue
}

\usepackage{amsmath,amssymb}
\usepackage{algorithm}
\usepackage{algpseudocode}
\usepackage{amsmath}      
\usepackage{amssymb}      
\usepackage{amsthm}       
\usepackage{tcolorbox}
\usepackage{booktabs}
\usepackage{bbding} 
\newcommand{\cmark}{\color{black}{\CheckmarkBold}}
\newcommand{\xmark}{\color{purple}{\XSolidBrush}}
\theoremstyle{plain}
\newtheorem{proposition}{Proposition}
\usepackage{amsmath, amsthm, amssymb}
\def\paperID{} 
\def\confName{CVPR}
\def\confYear{2026}

\title{Intrinsic Concept Extraction Based on Compositional Interpretability}

\author{
  Hanyu Shi\textsuperscript{1}\thanks{Equal contribution} \thanks{This work was completed during an internship at VIPSHOP}\quad
  Hong Tao\textsuperscript{2}\footnotemark[1] \quad
  Guoheng Huang\textsuperscript{1}\thanks{Corresponding authors} \quad
  Jianbin Jiang\textsuperscript{2} \\
  Xuhang Chen\textsuperscript{3}\footnotemark[3] \quad
  Chi-Man Pun\textsuperscript{4} \quad
  Shanhu Wang\textsuperscript{2}\footnotemark[3] \quad
  Pan Pan\textsuperscript{2} \\
  \textsuperscript{1}Guangdong University of Technology \quad
  \textsuperscript{2}VIPSHOP \quad
  \textsuperscript{3}Huizhou University \quad
  \textsuperscript{4}University of Macau
}

\begin{document}
\maketitle
\input{sec/0_abstract}    
\input{sec/1_intro}

\input{sec/2_relatedwork}
\input{sec/3_preliminaries}
\input{sec/4_model}
\input{sec/5_experiments}
\input{sec/6_conclusion}
{
    \small
    \bibliographystyle{ieeenat_fullname}
    \bibliography{main}
}

\end{document}

%% file: sec/0_abstract.tex
\begin{abstract}
Unsupervised Concept Extraction aims to extract concepts from a single image; however, existing methods suffer from the inability to extract composable intrinsic concepts. To address this, this paper introduces a new task called Compositional and Interpretable Intrinsic Concept Extraction (CI-ICE). The CI-ICE task aims to leverage diffusion-based text-to-image models to extract composable object-level and attribute-level concepts from a single image, such that the original concept can be reconstructed through the combination of these concepts. To achieve this goal, we propose a method called HyperExpress, which addresses the CI-ICE task through two core aspects. Specifically, first, we propose a concept learning approach that leverages the inherent hierarchical modeling capability of hyperbolic space to achieve accurate concept disentanglement while preserving the hierarchical structure and relational dependencies among concepts; second, we introduce a concept-wise optimization method that maps the concept embedding space to maintain complex inter-concept relationships while ensuring concept composability. Our method demonstrates outstanding performance in extracting compositionally interpretable intrinsic concepts from a single image.
\end{abstract}

%% file: sec/1_intro.tex
\section{Introduction}
\label{sec:intro}
\begin{figure}
    \centering
    \includegraphics[width=1\linewidth]{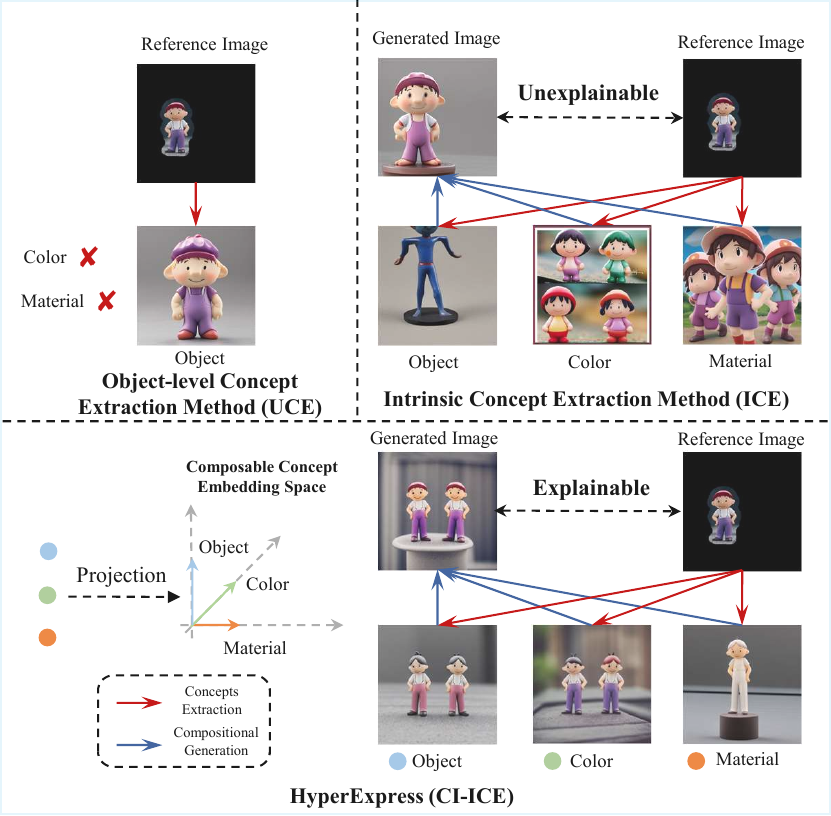}
    \caption{\textbf{The difference between unsupervised concept extraction (UCE) methods \cite{Stein2024,cendra2025ICE,AutoConcept} and the composable and interpretable intrinsic concept extraction method}. Object-level concept extraction method \cite{AutoConcept,hao2024conceptexpress} can only extract object-level concepts and is unable to extract attribute-level concepts such as color and material. Although the intrinsic concept extraction method \cite{cendra2025ICE} can extract both object-level concepts and attribute-level concepts, the concepts it extracts are not sufficiently close to the original concepts, and it fails to consider the compositionality of the embedding space; thus, it has poor interpretability. In contrast, HyperExpress considers the relationships between concepts when learning them. As a result, the extracted concepts are more aligned with the objects in the image. Additionally, it imposes compositional constraints on the concept embedding space, thereby enabling concept extraction and combination capabilities that are understandable to humans.
    }
    \label{fig:compared}
\end{figure}
Concept extraction \cite{Stein2024} aims to extract symbols with human-interpretable meanings from visual images and is often used to explain the behaviors of models. With the development of multimodal models \cite{zhang2026hallucination,yip2026c,yip2026sope}, in recent years, a variety of methods \cite{hao2024conceptexpress,10.1145/3610548.3618154,10.5555/3692070.3692963,cendra2025ICE,Stein2024} have explored the concept extraction capabilities of diffusion-based Text-to-Image (T2I) models. Concept extraction can be categorized into supervised concept extraction \cite{10.1145/3610548.3618154,10.5555/3692070.3692963} and unsupervised concept extraction \cite{hao2024conceptexpress,cendra2025ICE,Stein2024}.
Supervised concept extraction \cite{10.1145/3610548.3618154,10.5555/3692070.3692963} relies on external human-labeled knowledge for supervised learning, and this limitation poses significant obstacles to its practical applications. In contrast, unsupervised concept extraction \cite{hao2024conceptexpress,cendra2025ICE,Stein2024} aims to achieve concept extraction without relying on prior knowledge of concepts. Hao \etal~\cite{hao2024conceptexpress} proposed the Unsupervised Concept Extraction (UCE) task. UCE methods such as Break-A-Scene~\cite{10.1145/3610548.3618154}, ConceptExpress~\cite{hao2024conceptexpress}, and AutoConcept~\cite{AutoConcept} are designed to extract concepts from a single image; however, they can only extract object-level concepts. Cendra \etal~\cite{cendra2025ICE} proposed an intrinsic concept extraction method, ICE \cite{cendra2025ICE}, which is capable of extracting both object-level and attribute-level concepts from a single image.
Although these methods all perform well in the UCE \cite{hao2024conceptexpress} task, none of them consider the composability of concepts. This leads to poor interpretability, and there is uncertainty when using these concepts, which lack composability, to explain the original image content. This limits the ability to control and trust the model \cite{Stein2024}. Although Stein \etal \cite{Stein2024} proposed the CCE \cite{Stein2024} method for extracting composable concepts, it requires learning from multiple images containing the same concept, which also limits its application. Stein \etal~\cite{Stein2024} proposes that two composable concepts exhibit an approximately orthogonal relationship in the embedding space, and the effect of concepts is maximized when they possess composability. Current UCE methods \cite{cendra2025ICE,hao2024conceptexpress,AutoConcept} do not impose any constraints on the concept embedding space, which results in the concepts learned by these methods lacking composability and thus undermining their effectiveness.

To address the problem of extracting composable object-level concepts and attribute-level concepts from a single image, we propose a new task called Compositional and Interpretable Intrinsic Concept Extraction (CI-ICE). CI-ICE aims to leverage diffusion-based Text-to-Image (T2I) \cite{Rombach2021HighResolutionIS} models to extract object-level concepts and their corresponding attribute-level concepts from a single reference image. Moreover, these extracted concepts can be combined to form a complete sample, which is used to explain the complex concepts in the original reference image. In \cref{fig:compared}, we illustrate the distinctions between the method based on CI-ICE and existing concept extraction approaches. Current concept extraction methods place greater emphasis on concept disentanglement while overlooking their composability, thus failing to adequately accomplish the CI-ICE task. The CI-ICE task faces two challenges: 
\begin{itemize}
    \item The accurate disentanglement of object-level concepts and attribute-level concepts: there exist obvious hierarchical structures and associative relationships between these two types of concepts.
    \item The learned concept embedding space is required to have composability: only composable concepts can better explain the original complex concepts in the reference image.
\end{itemize}

To address the two challenges of the CI-ICE task, we propose the HyperExpress method. The HyperExpress method tackles the two challenges of CI-ICE from two aspects: concept learning and concept optimization. In terms of concept learning, we propose a Hyperbolic Contrastive Learning module and a Hyperbolic Entailment Learning module. The Hyperbolic Contrastive Learning module leverages the inherent hierarchical modeling capability of hyperbolic space. In this design, object-level concepts and attribute-level concepts are positioned at different locations in the space, distinguishing object-level concepts from attribute-level concepts within complex concepts. The Hyperbolic Entailment Learning module, on the other hand, establishes relational dependencies between object-level and attribute-level concepts based on the hyperbolic entailment cone. In terms of concept optimization, we aim to achieve concept composability while preserving the hierarchical structure and relational dependencies among concepts through a Horosphere Projection module. As shown in \cref{fig:compared}, compared to existing concept extraction methods, our approach not only accurately disentangles object-level and attribute-level concepts but also ensures concept composability through the optimization of the concept embedding space. Experimental validation demonstrates that the HyperExpress method exhibits promising potential in mining composable visual concepts. Our contributions are summarized below:
\begin{itemize}
    \item We propose the task of Compositionally Interpretable Intrinsic Concept Extraction (CI-ICE), aiming to address the issue that existing unsupervised concept extraction (UCE) methods fail to extract composable intrinsic concepts.
    \item We propose the HyperExpress method, which aims to extract composable intrinsic concepts from a single image to accomplish the CI-ICE task.
    \item We evaluated the effectiveness of the HyperExpress method on the UCE benchmark, and the experiments demonstrate that the HyperExpress method is a highly promising solution for extracting composable intrinsic concepts.
\end{itemize}

%% file: sec/2_relatedwork.tex
\section{Related Work}
\label{sec:related work}

\subsection{Generative Concept Learning}
Generative Concept Learning aims to decompose a complex visual concept into simple, basic elements. Some methods \cite{10.5555/3666122.3667440,citation-key,10.1145/3592133,hao2023vicoplugandplayvisualcondition,Jia2023TamingEF,Kumari2022MultiConceptCO,Li2023BLIPDiffusionPS,Ma2023UnifiedML,10.5555/3666122.3669594,10204880,Shi2023InstantBoothPT,Tewel2023KeyLockedRO,10377571} extract concepts from multiple images that contain the same concept. For example, the Textual Inversion \cite{citation-key} method represents a certain concept by learning an embedding vector, but it can only learn a single concept; Liu \etal \cite{10377938} extended the Textual Inversion method to extract multiple concepts. These methods all rely on multiple images representing the same concept, and this limitation undermines their practicality. Although methods such as Break-A-Scene \cite{10.1145/3610548.3618154}, MCPL \cite{10.5555/3692070.3692963}, and DisenDiff \cite{10656657} can extract concepts from a single image, they depend on manually provided prior knowledge. ConceptExpress \cite{hao2024conceptexpress} proposed the task of unsupervised concept extraction, which aims to extract concepts from a single image without relying on manually provided prior knowledge, but it cannot distinguish between object-level concepts and attribute-level concepts. Inspiration Tree \cite{10.1145/3618315} forces the model to learn different concepts by decomposing tokens, but it uses structured guidance. ICE \cite{cendra2025ICE} can structurally extract object-level concepts and attribute-level concepts from a single image, yet it fails to consider the associative relationships between object-level concepts and attribute-level concepts.

None of these methods take into account the compositionality and interpretability of the extracted concepts. This makes the process of decomposing a complex visual concept into multiple simple visual concepts irreversible, resulting in insufficient interpretability of these methods and thus weakening people's ability to control and trust the model. Although CCE \cite{Stein2024} considers the compositionality of the concept embedding space, it still needs to extract concepts from multiple images containing the same concept. The HyperExpress method we propose not only extracts object-level concepts and attribute-level concepts in a structured manner and learns the associative relationships between concepts but also imposes compositional constraints on the embedding space, maximizing the realization of reversible decomposition from a complex visual concept to multiple simple visual concepts.

\subsection{Hyperbolic learning}
Euclidean space has been used for representation learning \cite{10483696}. However, computer vision data often exhibits a highly non-Euclidean underlying geometric structure, in which case Euclidean embedding may not be the optimal choice \cite{7974879,HyperNN}. Hyperbolic space inherently has the ability to represent hierarchical structures with minimal distortion \cite{10.5555/3495724.3496555,Kim2020MixCoMC,10.1007/978-3-030-01246-5_45,Liu2022EnhancingHG}, which makes it possible to learn hierarchical visual concept embeddings in hyperbolic space. Recent works \cite{Desai2023Hyperbolic,Ibrahimi2024IntriguingPO,Pal2024CompositionalEL} have demonstrated the potential of hyperbolic learning in learning cross-modal hierarchical embeddings in vision-language models, among which the study by Desai \etal \cite{Desai2023Hyperbolic} enforces the construction of entailment structures across modalities. Drawing inspiration from these methods, we establish hierarchical structures and entailment relationships between object-level concepts and attribute-level concepts, which can be used to realize the decomposition of complex visual concepts.

\subsection{Compositionality in Concept Extraction}
Existing studies \cite{wang2024conceptalgebrascorebasedtextcontrolled,Kwon2022DiffusionMA} have shown that compositionality can be used to control the behavior of generative models. Current concept extraction models focus on obtaining disentangled representations of concepts. ``Disentanglement" focuses on how to distinguish different concepts, while ``compositionality" focuses on the results produced when different concepts are combined with each other. Nevertheless, there is no inherent correlation between disentanglement and compositionality \cite{10.5555/3600270.3602088}. Stein \etal \cite{Stein2024} discussed the compositionality of unsupervised concept extraction methods and proposed that the concepts extracted by current unsupervised concept extraction methods lack compositionality, as no constraints are imposed on the concept embedding space. Building on the work of CCE \cite{Stein2024}, we further consider emphasizing not only the disentanglement capability of concept extraction models but also their compositional generation capability.

%% file: sec/3_preliminaries.tex
\section{Preliminaries}
\subsection{Diffusion Models}
We use diffusion-based Text-to-Image (T2I) models \cite{diffusion1,Nichol2021GLIDETP,Ramesh2022HierarchicalTI,Rombach2021HighResolutionIS,10.5555/3600270.3602913,Song2020DenoisingDI} for Compositional and Interpretable Intrinsic Concept Extraction (CI-ICE). A diffusion model gradually adds noise to data to convert the data into a random distribution and then reconstructs the original data from the noise through a reverse process. In the forward process, the diffusion model adds Gaussian noise to the original sample $z_0$ to obtain $z_t$:
\begin{equation}
    q\left(z_{t} \mid z_{0}\right)=\mathcal{N}\left(z_{t} ; \sqrt{\bar{\alpha}_{t}} z_{0},\left(1-\bar{\alpha}_{t}\right) \mathbf{I}\right),
\end{equation}
where $\bar{\alpha}_{t} = \prod_{i=1}^{t} \alpha_{i}$
, $\alpha_{i}$ represents the noise schedule. During the denoising process, the denoising network predicts the noise at the current time step based on the noisy image $z_t$ at time step $t$. This process is achieved by minimizing the denoising loss:
\begin{equation}
    \mathcal{L}_{\text{recon}} = \mathbb{E}_{\mathbf{x}_{0}, \epsilon, t}\left[ \left\| c - c_{\theta}\left( \mathbf{x}_{t}, t, \mathcal{E}(\mathbf{p}) \right) \right\|_{2}^{2} \right].
\end{equation}

\subsection{Hyperbolic Geometry}
To learn the hierarchical relationships and associative relationships between concepts in greater detail, our method operates in hyperbolic space. Hyperbolic space is a Riemannian manifold with constant negative curvature, which ensures that in hyperbolic space, the embedding positions of concepts with greater differences are farther apart, while concepts with smaller differences are closer to each other—this is difficult to achieve in Euclidean space. Specifically, we select the Poincar\'{e} ball \cite{HyperNN} for concept learning. The Poincar\'{e} ball \cite{HyperNN} $(\mathbb{B}^n,g^\mathbb{B})$ is defined by
the manifold $\mathbb{B}^n=\{x\in\mathbb{R}^n:\|x\|<1\}$ equipped with the Riemannian metric $g_x^\mathbb{B}=\lambda_x^2g^E$. Similar to concept extraction models based on Euclidean space, we use hyperbolic distance to measure the similarity between two concepts, which is defined as:
\begin{equation}
d_\mathbb{D}(x,y)=\cosh^{-1}\left(1+2\frac{\|x-y\|^2}{(1-\|x\|^2)(1-\|y\|^2)}\right) \label{eq:d}.
\end{equation}

In Euclidean space, we often use orthogonal relationships to represent the composability of two concepts, and two mutually orthogonal concepts can be combined. This first requires defining the directionality of the embedding space. In hyperbolic space, ideal points are used to represent directions, and they refer to the points on the unit sphere $\mathbb{S}_{\infty}^{d-1}=\{\|x\|=1\}$. A geodesic is analogous to a straight line in Euclidean space. We can define a submanifold $M \subset \mathbb{H}^{d}$ such that: for any $x, y \in M$, the geodesic connecting $x$ and $y$ is contained in $M$; such a submanifold is called a geodesic submanifold. 
Furthermore, given a set of points $S \in \mathbb{H}^d$, the smallest geodesic submanifold in the hyperbolic space $\mathbb{H}^d$ that contains $S$ is called the geodesic hull of $S$, denoted as $GH(S)$. In the Poincar\'{e} ball model, the level set of the Busemann function is called the horosphere centered at $p$. The intrinsic curvature of a horosphere is zero; therefore, it also possesses many properties of planes in Euclidean space.
 
\subsection{Concepts and Compositionality}
In our research, a concept is defined as a feature or attribute in an image that is clearly distinguishable and understandable to humans \cite{Stein2024}, such as objects, colors, and the materials of objects in the image. We use a set of tokens $\mathcal{T}$ to refer to these concepts, along with their corresponding embedding vectors $\mathcal{V}$. Furthermore, we define items, characters, and animals \etc in the image as object-level concepts, and attributes such as color, shape, and size as attribute-level concepts. The compositionality of concepts is defined in \cref{propositionR}.
\begin{proposition}
\label{propositionR}
For concept tokens $[V_i],[V_j] \in \mathcal{T}$, the concept representation $R:\mathcal{T} \to \mathcal{V}$ is considered compositional if there exist positive weights $w_i,w_i \in \mathbb{R}^+$ such that:
\begin{equation}
    R([V_i]\cup [V_j])=w_iR([V_i])+w_jR([V_j]).
\end{equation}
\end{proposition}

%% file: sec/4_model.tex
\section{Method}
\begin{figure*}
    \centering
    \includegraphics[width=0.9\linewidth]{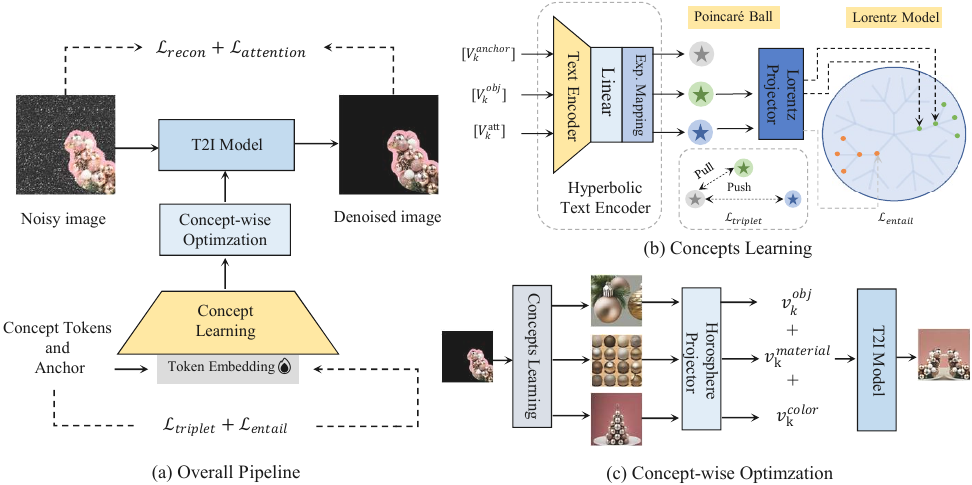}
    \caption{\textbf{The Proposed Method and Its Components}. \textbf{(a) The overall structure of HyperExpress}: It addresses the CI-ICE task from two aspects: concept learning and concept-wise optimization. \textbf{(b) Concept learning}: It leverages triplet loss $\mathcal{L}_{triplet}$ and hyperbolic entailment loss $\mathcal{L}_{entail}$ to learn the hierarchical structure and associative relationships between object-level concepts and attribute-level concepts. \textbf{(c) Concept-wise Optimization}: It uses Horosphere Projector (HP) to constrain the concept space, thereby ensuring the compositionality of concepts.}
    \label{fig:overview}
\end{figure*}

We aim to learn a set of tokens that can refer to object-level concepts and attribute-level concepts from a single image, and these concepts need to be compositional. Specifically, given an image $\mathcal{I}$ containing $N$ objects where each object has $M$ attributes, we leverage diffusion-based Text-to-Image (T2I) models \cite{diffusion1,Nichol2021GLIDETP,Ramesh2022HierarchicalTI,Rombach2021HighResolutionIS,10.5555/3600270.3602913,Song2020DenoisingDI} to mine a set of concept tokens $\mathcal{T}=\{[V_i]\}_{i=1}^{\text{(M+1)} \cdot \text{N}}$ and their embedding vectors $\mathcal{V}=\{v_i\}_{i=1}^{\text{(M+1)} \cdot \text{N}}$. These tokens $\mathcal{T}$ and embedding vectors $\mathcal{V}$ can capture specific concepts in the image, such as the category, color, and material of objects within the image $\mathcal{I}$. Furthermore, we impose constraints on the embedding space to ensure it satisfies \cref{propositionR}, thereby enabling the extraction of concepts with compositional interpretability. To this end, we propose the HyperExpress method, which consists of a \textbf{Concept Learning} approach and a \textbf{Concept-wise Optimization} approach. \Cref{fig:overview} illustrates the overall framework of the HyperExpress method.

\subsection{Concept Learning}
The first challenge of the CI-ICE task is how to accurately disentangle complex concepts, and this stage aims to learn a set of concept tokens $\mathcal{T}=\{\mathcal{T}^{obj},\mathcal{T}^{att}\}$ and their embedding vectors $\mathcal{V}=\{\mathcal{V}^{obj},\mathcal{V}^{att}\}$, so as to disentangle the complex concepts in the image. For an image containing $N$ objects, where each object has $M$ attributes, we have $\mathcal{T}^{obj}=\{[V_k^{obj}]\}_{k=1}^\text{N}$, $\mathcal{V}^{obj}=\{v_k^{obj}\}_{k=1}^\text{N}$, and $\mathcal{T}^{att}=\{[V_k^{att}]\}_{k=1}^{\text{N}\cdot\text{M}}$, $\mathcal{V}^{att}=\{v_k^{att}\}_{k=1}^{\text{N}\cdot\text{M}}$. There are obvious hierarchical and associative relationships between object-level concepts and attribute-level concepts; methods based on Euclidean space cannot accurately capture such relationships. However, hyperbolic space inherently possesses strong hierarchical modeling capabilities, which is why we conduct the learning process in hyperbolic space. Since our focus is on disentangling complex concepts in the embedding space, we first adopt the method from the first stage of ICE \cite{cendra2025ICE} to locate the main objects in the image and their semantic categories, and finally obtain the masks $\mathcal{M}=\{M_i\}_{i=1}^\text{N}$ of the main objects and their corresponding text descriptions $\mathcal{T}^{anchor}=\{[V_k^{anchor}]\}_{k=1}^\text{N}$. Subsequently, the Hyperbolic Contrastive Learning (HCL) module is used to learn the hierarchical structure between concepts, and the Hyperbolic Entailment Learning (HEL) module is used to learn the entailment relationships between concepts.

\noindent \textbf{Hyperbolic Contrastive Learning Module}. The HCL module aims to distinguish between object-level concepts and attribute-level concepts by leveraging the hierarchical modeling capability of hyperbolic space.

First, we define the hyperbolic text encoder. For the newly added tokens of each main object, we first encode them using the CLIP \cite{Radford2021LearningTV} model, and then map the text embeddings to the Poincar\'{e} ball \cite{HyperNN,Peng2021HyperbolicDN} through the exponential map \cite{HyperNN,Peng2021HyperbolicDN}. We add a weight $W$ to the exponential map to learn the mapping relationship from the standard text encoder space to the tangent space, as follows:
\begin{equation}
\mathcal{E}_h(x) = \exp_0(W\cdot\mathcal{E}_s(x)),\label{eq:eh}
\end{equation}
where $\mathcal{E}_h(\cdot)$ is the hyperbolic text encoder and $\mathcal{E}_s(\cdot)$ is the CLIP \cite{Radford2021LearningTV} text encoder. Exponential map $\exp_0(\cdot)$ refers to the exponential operation in the Poincar\'{e} ball \cite{HyperNN,Peng2021HyperbolicDN}, and its calculation is as follows:
\begin{equation}
    \exp_{0}(\mathbf{x}) = \tanh\left( \frac{\|\mathbf{x}\|}{2} \right) \cdot \frac{\mathbf{x}}{\|\mathbf{x}\|},
    \label{eq:exp}
\end{equation}
therefore, we have $\mathcal{V}^{obj}=\{\mathcal{E}_h([V_k^{obj}])\}_{k=1}^{\text{N}}$ and $\mathcal{V}^{att}=\{\mathcal{E}_h([V_k^{att}])\}_{k=1}^{\text{N}\cdot\text{M}}$.

Inspired by \cite{cendra2025ICE}, we use the hyperbolic triplet loss to learn to distinguish between concepts. 

The first step is to distinguish between object-level concepts and attribute-level concepts, as follows:
\begin{equation}
    \mathcal{L}_{\text{triplet,k}}^{obj} = \max\Big(0, d_{\mathbb{D}}(v_k^{\text{anchor}}, v_k^{\text{obj}}) - d_{\mathbb{D}}(v_k^{\text{anchor}}, v_k^{\text{att}}) + \gamma \Big),
\end{equation}
where $v_k^{\text{anchor}}=\mathcal{E}_h([V_k^{anchor}])$ and $\gamma$ are the margin parameters. We use $v_k^{\text{anchor}}$ to initialize $v_k^{\text{obj}}$.

Next is the distinction between different attribute-level concepts, as follows:
\begin{equation}
    \mathcal{L}_{\text{triplet,k}}^{att} = \max\Big(0, d_{\mathbb{D}}(v_k^{\text{intrinsic}}, v_k^{\text{att}}) - d_{\mathbb{D}}(v_k^{\text{intrinsic}}, v_j^{\text{att}}) + \gamma \Big),
\end{equation}
where $v_k^{\text{intrinsic}}=\mathcal{E}_h([V_k^{intrinsic}])$ and $[V_k^{intrinsic}]$ are set to $[V_k^{intrinsic}]$: ``a $intrinsic_k$ concept" where $intrinsic_k$ represents a specific attribute such as color.
In contrast to ICE \cite{cendra2025ICE}, we distinguish these concepts in hyperbolic space, which makes the concepts we have learned more hierarchical; however, the associative relationships between concepts are not considered. To address this issue, we propose the Hyperbolic Entailment Learning (HEL) module.

\noindent \textbf{Hyperbolic Entailment Learning Module}. Inspired by \cite{pmlr-v80-ganea18a}, we define the associative relationship between concepts as follows: if concept $i$ entails concept $j$, then their tokens $[V_i],[V_j]$ and embeddings $v_i=\mathcal{E}_h([V_i]),v_j=\mathcal{E}_h([V_j])$ satisfy the following formula:
\begin{equation}
    v_i \in \mathcal{S}_{j} \Longleftrightarrow \theta(v_i,v_j) \leq \omega(v_i),\label{eq:yh}
\end{equation}
where $w(\cdot)$ denotes the entailment cone radius, $\theta(i,j)$ denotes the spatial angle, and $\mathcal{S}_{i}$ denotes the entailment cone of concept $i$. Since both $w$ and $\theta$ have explicit formula definitions in the Lorentz model \cite{Peng2021HyperbolicDN}, and their calculation is simpler compared to the Poincar\'{e} model \cite{HyperNN}, we perform the computation in the Lorentz model \cite{Peng2021HyperbolicDN}. That uses the following formula \cite{Peng2021HyperbolicDN}:
\begin{multline}
x = (x_0, \dots, x_n) \in \mathbb{B}^n \Leftrightarrow \\
\left( \frac{1 + \|x\|^2}{1 - \|x\|^2}, \frac{2x_1}{1 - \|x\|^2}, \dots, \frac{2x_n}{1 - \|x\|^2} \right) \in \mathbb{L}^n,
\end{multline}
where, $\mathbb{B}^n$ and $\mathbb{L}^n$ are the $n$-dimensional Poincar\'{e} ball \cite{HyperNN,Peng2021HyperbolicDN} and Lorentz model \cite{Peng2021HyperbolicDN} and $\|x\|^2=x_0^2 + \dots + x_n^2$.

\begin{figure}
    \centering
    \includegraphics[width=0.9\linewidth]{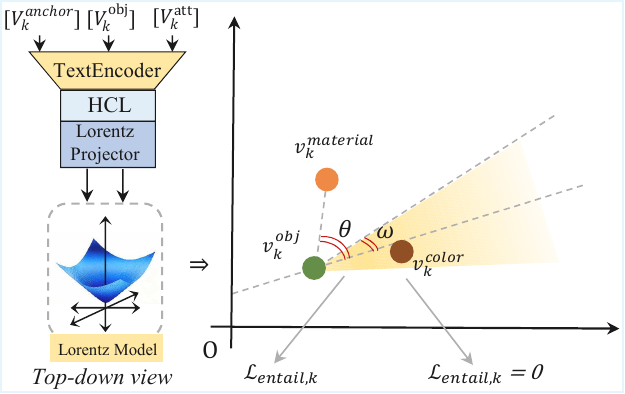}
    \caption{\textbf{Explanation of the proposed HEL module}. Unlike the HCL module, the hyperbolic entailment loss is computed within the Lorentz model. If the object-level concept ($v_k^{obj}$) and attribute-level concept ($v_k^{color}$ and $v_k^{material}$) satisfy the condition in \cref{eq:yh}, the entailment loss will be $0$; otherwise, the corresponding entailment loss will be calculated.}
    \label{fig:EM}
\end{figure}

In the Lorentz model \cite{Peng2021HyperbolicDN}, the definitions of the entailment cone radius $\omega(\cdot)$ are as follows \cite{Desai2023Hyperbolic,Le2019InferringCH,Poppi2025HyperbolicSV}:
\begin{equation}
    \omega(x) = \sin^{-1}\left( \frac{2K}{\sqrt{\kappa} \|x\|} \right),
\end{equation}
and the spatial angle $\theta(x,y)$ is as follows:
\begin{equation}
\theta(x, y)=\cos^{-1}\left(\frac{x_{0} + y_{0} \cdot \kappa \cdot \langle x, y \rangle_{\mathcal{L}}}{\|\tilde{y}\| \cdot \sqrt{ \left( \kappa \cdot \langle x, y \rangle_{\mathcal{L}} \right)^{2} - 1 }}\right),
\end{equation}
where $\kappa$ denotes the curvature, and $\langle x, y \rangle_{\mathcal{L}}$ represents the Lorentz inner product.
Next, based on \cref{eq:yh}, we set up an entailment loss. The entailment loss between object-level concepts and attribute-level concepts is shown as follows:
\begin{equation}
    \mathcal{L}_{\text{entail,k}} = \max \Bigl(0, \cos(\omega(v_k^{obj})) - \cos(\theta(v_k^{obj},v_k^{att}))\Bigr),
\end{equation}
where $\mathcal{L}_{\text{entail,k}}$ denotes the entailment loss of the $k$-th object concept in the image. As shown in \cref{fig:EM}, this means that the attributes of this object should fall within the entailment cone of the object. By constraining the embedding positions of object-level concepts and attribute-level concepts, we establish the associative relationship between them and achieve accurate disentanglement of complex concepts. Next, we impose further constraints on the concept embedding vectors to ensure their composability.

\subsection{Concept-wise Optimization}
Another core challenge faced by CI-ICE lies in the composability of the learned concept embedding vectors. Currently, mainstream unsupervised concept extraction methods \cite{cendra2025ICE,hao2024conceptexpress} have failed to address this issue effectively, and the fundamental reason is that these methods lack an explicit constraint mechanism on the concept embedding space. Although the CCE \cite{Stein2024} method proposed by Stein \etal incorporates constraints, its modeling in Euclidean space makes it difficult for the model to effectively learn the hierarchical structures and associative relationships between concepts. In view of this, inspired by \cite{chami2021horopca}, this study proposes a concept composability Horosphere Projection (HP) module suitable for hyperbolic space. The core goal of HP is to not only maintain the complex intrinsic relationships between concepts in hyperbolic space but also to achieve the composability of concept embedding vectors, thereby solving the aforementioned key issue in a targeted manner.

\noindent \textbf{Horosphere Projection Module}. Given $n$ ideal points $\{p_1,p_2,\dots,p_n\}$ and one base point $b$, our goal is to learn a mapping from the concept embedding space to the composable submanifold.
\begin{equation}
\pi_{b,p_{1},\ldots,p_{n}}^{\mathbb{H}} : 
v \mapsto M \cap S(p_{1},v) \cap \cdots \cap S(p_{n},v),
\end{equation}
where $M:=GH(b,p_1,\dots,p_n)$ and $S(p_i,v)$ denote the horosphere centered at $p_i$ and passing through the concept embedding point $v$. 

We train our HP on the anchors of concepts. Since anchors have been proven to be composable in the embedding space \cite{Stein2024}, our goal is to enable the newly added anchors to possess the composability of the original semantic concepts. The training objective is to find $n$ geodesic directions such that the variance of the data after horosphere projection is maximized, as shown in the following formula:
\begin{equation}
\begin{split}
    &p_{1} = \underset{p \in \mathbb{S}_{\infty}^{d-1}}{\operatorname{argmax}} \sigma_{\mathbb{H}}^{2}\left(\pi_{b, p}^{\mathbb{H}}(S)\right), \\
&p_{k+1} = \underset{p \in \mathbb{S}_{\infty}^{d-1}}{\operatorname{argmax}} \sigma_{\mathbb{H}}^{2}\left(\pi_{b, p_{1}, \ldots, p_{n}, p}^{\mathbb{H}}(S)\right), \\
&\sigma_{\mathbb{H}}^{2}(S)=\frac{1}{n^{2}} \sum_{x, y \in S} d_{\mathbb{H}}(x, y)^{2}.
\end{split}
\end{equation}

Algorithm 1 in Appendix 7 illustrates the calculation process of the HP module.

\begin{figure}
    \centering
    \includegraphics[width=0.8\linewidth]{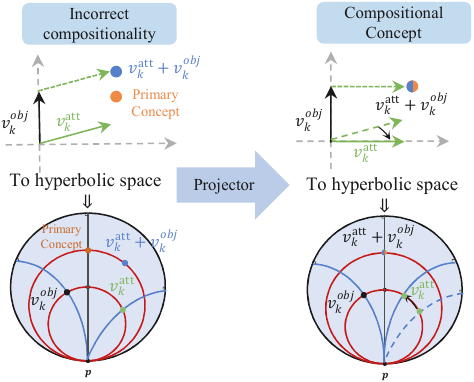}
    \caption{\textbf{Horosphere Projection Module and an illustration of concept compositionality}. If no constraints are applied to the concept embedding space, it may lead to incorrect composition of concepts. For instance, using $v_k^{att}+v_k^{obj}$ alone cannot reconstruct the primary concept. To address this, we reproject the embedding space to ensure composability between concepts.}
    \label{fig:Compositional}
\end{figure}

As shown in \cref{fig:Compositional}, HP essentially uses anchors to find $n$ geodesic directions and adjusts the newly added Token vectors to these $n$ geodesic directions. The learned submanifold inherits Euclidean properties from the zero-curvature horosphere, allowing native operations like vector addition \cite{chami2021horopca}. Therefore, after finding the n geodesic directions, we can complete the rotation operation using an orthogonal matrix $Q$.

The HP module is designed based on the isometry of horospherical projection, which endows it with two advantages. The isometry of horospherical projection is defined as \cref{proposition1}.
\begin{proposition}
\label{proposition1}
For any $x \in \mathbb{H}^{d}$, if $y \in \operatorname{GH}(x, ,p_1,\dots,p_n)$, then:
\begin{equation}
    d_{\mathbb{H}}(\pi_{b,p_1,\dots,p_n}^{\mathbb{H}}(x), \pi_{b,,p_1,\dots,p_n}^{\mathbb{H}}(y)) = d_{\mathbb{H}}(x, y).
\end{equation}
\end{proposition}

The one advantage is that the HP module does not disrupt the hierarchical structures and associative relationships between concepts. The complex relationships between concepts are mainly determined by their distances: the stronger the association between two concepts, the closer their distance should be. Since the HP module satisfies the \cref{proposition1}, it does not alter the distance between two concepts before projection; thus, it does not affect concept disentanglement. We have proven the \cref{proposition1} in Appendix 1. Another advantage of the HP module is that, through its rotation operation, we project the newly added concepts into a composable concept space. This space can satisfy \cref{propositionR}, thereby enabling composability. We have provided the corresponding proof process in Appendix 2.

\subsection{Overall training objective}
The method we proposed includes multiple loss terms, and the total loss is as follows:

\begin{equation}
    \mathcal{L} = \mathcal{L}_{\text{recon}} + \lambda_{\text{triplet}} \cdot \mathcal{L}_{\text{triplet}} \\
    + \lambda_{\textit{attention}} \cdot \mathcal{L}_{\textit{attention}} + \lambda_{\text{entail}} \cdot \mathcal{L}_{\textit{entail}}.
\end{equation}

Since the concept learning process involves learning the embeddings of object-level concepts and the embeddings of attribute-level concepts, we have two types of triplet losses:
\begin{equation}
    \mathcal{L}_{\text{triplet}}^{\text{obj}} = \sum_{k} \mathcal{L}_{\text{triplet,k}}^{\text{obj}}, \quad \mathcal{L}_{\text{triplet}}^{\text{att}} = \sum_{k} \mathcal{L}_{\text{triplet,k}}^{\text{att}}.
\end{equation}
Similarly, we also have entailment loss:
\begin{equation}
    \mathcal{L}_{\text{entail}} = \sum_{k} \mathcal{L}_{\text{entail,k}}.
\end{equation}
The attention loss $\mathcal{L}_{\textit{attention}}$ adopts the Wasserstein loss \cite{hao2024conceptexpress}, which can align the attention of the T2I model with the masked regions, thereby reducing the impact of object-irrelevant features:
\begin{equation}
    \mathcal{L}_{\text{attention}} = \mathcal{W}(A_i,M_i),
\end{equation}
where $\mathcal{W}$ is Wasserstein distance and $A_i$, $M_i$ are the attention region and mask of the i-th concept, respectively.

%% file: sec/5_experiments.tex
\section{Experiments}
\begin{figure*}
    \centering
    \includegraphics[width=\linewidth]{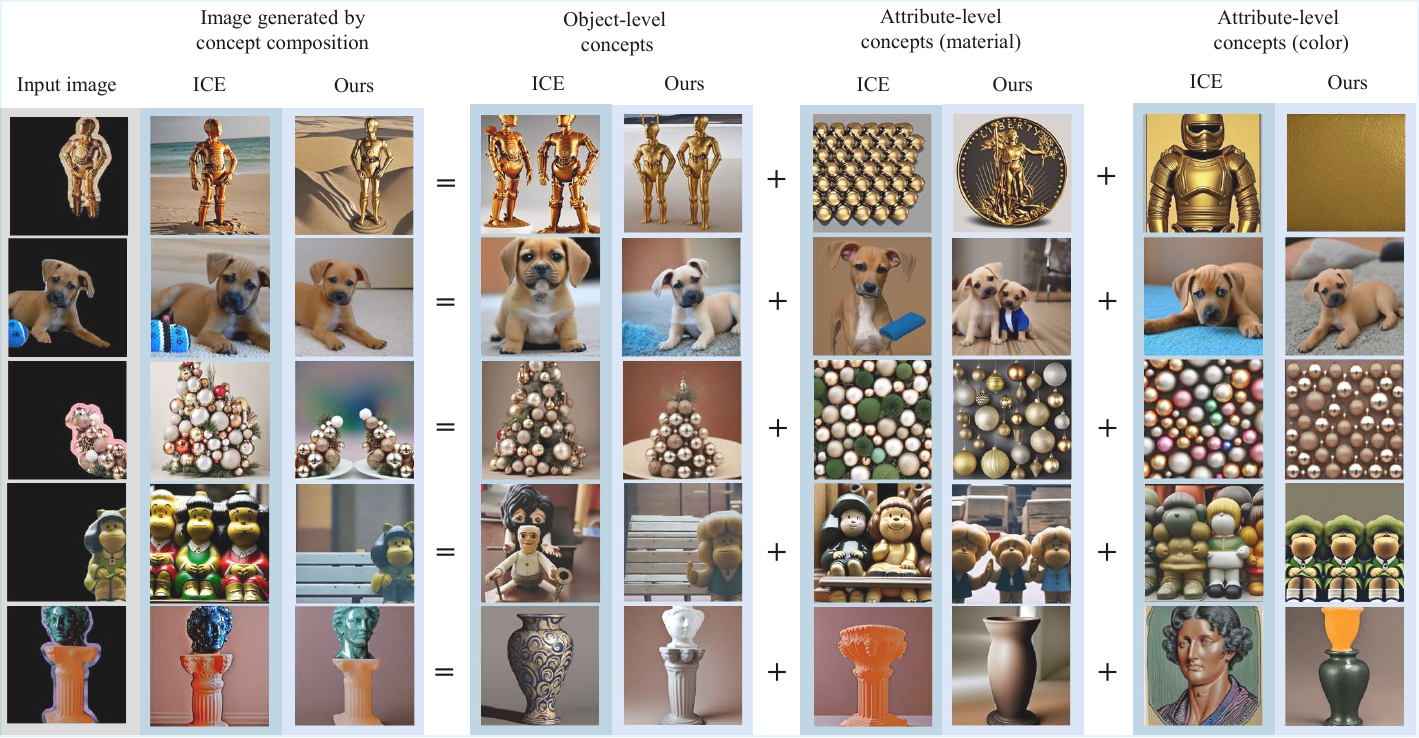}
    \caption{\textbf{Comparison of Qualitative Results Between HyperExpress and ICE \cite{cendra2025ICE}}. This comparison includes two processes: one is extracting object-level concepts and further attribute-level concepts from a single image, and the other is performing compositional reconstruction using the extracted concepts. }
    \label{fig:result}
\end{figure*}

\subsection{Baseline and Evaluation Metrics}
We evaluated HyperExpress on the Unsupervised Concept Extraction Benchmark (UCEBench) \cite{hao2024conceptexpress} and the Intrinsic Concept Benchmark (ICBench) \cite{cendra2025ICE}. Specifically, on the UCEBench \cite{hao2024conceptexpress}, we compared it with state-of-the-art UCE models \cite{hao2024conceptexpress,AutoConcept,10.1145/3610548.3618154,cendra2025ICE} using the identity similarity $\text{SIM}^I$, compositional similarity $\text{SIM}^C$, and Top-$k$ accuracy $\text{ACC}^k$ metrics. On the ICBench \cite{cendra2025ICE}, we conducted comparisons using the $\text{SIM}^{T-T}$ and $\text{SIM}^{T-V}$ metrics. We conducted training and testing on the $D1$ dataset \cite{cendra2025ICE} by referring to the approach of ICE \cite{cendra2025ICE}.
Among these metrics, $\text{SIM}^I$ measures the reconstruction accuracy of individual concepts; $\text{SIM}^C$evaluates the overall consistency of generated images based on extracted concepts; $\text{ACC}^k$ assesses the concept disentanglement capability of models; $\text{SIM}^{T-T}$ calculates the similarity between concepts described by GPT \cite{openai2024gpt4ocard} and learned tokens; $\text{SIM}^{T-V}$ computes the similarity between concepts described by GPT \cite{openai2024gpt4ocard} and images generated using the learned concepts. We present more experimental details and the calculation methods of evaluation metrics in Appendix 5.

\subsection{Quantitative results}
\begin{table}[ht]
\centering
\caption{Performance comparison on UCEBench \cite{hao2024conceptexpress}.}
\label{tab:clip_performance}
\resizebox{0.45\textwidth}{!}{%
\begin{tabular}{@{}ccccc@{}}
    \toprule
    \multicolumn{1}{c}{Method} & $\text{SIM}^I$ (\%) & $\text{SIM}^C$ (\%) & $\text{ACC}^1$ (\%) & $\text{ACC}^3$ (\%) \\ 
    \midrule
    Break-A-scene \cite{10.1145/3610548.3618154} & $0.627$ & $0.773$ & $0.174$ & $0.282$ \\
    ConceptExpress \cite{hao2024conceptexpress} & $0.689$ & $0.784$ & $0.263$ & $0.385$ \\
    AutoConcept \cite{AutoConcept} & $0.690$ & $0.770$ & $\underline{0.350}$ & $\underline{0.520}$ \\ 
    ICE \cite{cendra2025ICE} & $\textbf{0.738}$ & $\textbf{0.822}$ & $0.325$ & $0.518$ \\ 
    \midrule
    \rowcolor{blue!5!white} \textbf{HyperExpress (Ours)} & $\underline{0.699}$ & $\underline{0.786}$ & $\textbf{0.504}$ & $\textbf{0.736}$ \\
    \bottomrule
\end{tabular}%
}
\end{table}

\begin{table}[ht]
\centering
\caption{Performance comparison on ICBench \cite{cendra2025ICE}.}
\label{tab:dino_performance}
\resizebox{0.45\textwidth}{!}{%
\begin{tabular}{@{}ccccccc@{}}
    \toprule
    \multicolumn{1}{c}{Method} & $\text{SIM}^{T-T}_{object}$ & $\text{SIM}^{T-T}_{material}$ & $\text{SIM}^{T-T}_{color}$ & $\text{SIM}^{T-V}_{object}$ &
    $\text{SIM}^{T-V}_{material}$ & $\text{SIM}^{V-T}_{color}$ \\ 
    \midrule
    ICE \cite{cendra2025ICE} & $0.249$ & $0.101$ & $0.093$ & $0.264$ &$0.208$ &$0.215$ \\
    \midrule
    \rowcolor{blue!5!white} \textbf{HyperExpress (Ours)} & $\textbf{0.280}$ & $\textbf{0.115}$ & $\textbf{0.098}$ & $\textbf{0.305}$ & $\textbf{0.211}$ & $\textbf{0.222}$\\
    \bottomrule
\end{tabular}%
}
\end{table}

As shown in \cref{tab:clip_performance,tab:dino_performance}, HyperExpress demonstrates competitive performance compared with existing UCE models \cite{10.1145/3610548.3618154,cendra2025ICE,AutoConcept,hao2024conceptexpress}. However, it is worth noting that ICE \cite{cendra2025ICE} sacrifices the interpretability of concept composition. The paths generated by its concept composition are difficult to understand, which is not conducive to people's control over the model. Although the HyperExpress model we proposed loses some performance, it gains easily interpretable concept composition paths and a composable concept space.

\subsection{Qualitative results}
The CI-ICE task requires extracting intrinsic concepts with compositionality from images. Among existing UCE methods, only ICE \cite{cendra2025ICE} can extract intrinsic concepts; therefore, we conduct a qualitative comparison between HyperExpress and ICE \cite{cendra2025ICE} to evaluate the models’ concept extraction and combination capabilities, with the results shown in \cref{fig:result}.
In concept extraction, HyperExpress distinguishes itself from ICE \cite{cendra2025ICE} by learning associative relationships between concepts, leading to the extraction of specific, concrete object concepts from images. This specificity enhances the interpretability of the compositional process. Consequently, in compositional reconstruction, HyperExpress generates more interpretable pathways than ICE. For instance, as shown in \cref{fig:result}, it logically combines the concepts of ``robot," ``metal," and ``gold" to form the complex concept ``a golden robot made of metal." This clarity stems from learning concept associations and applying constraints to the concept embedding space, whereas ICE \cite{cendra2025ICE}'s compositional results remain difficult to interpret. We also provide the user study in Appendix 6.

\subsection{Model component analysis}
To verify the effectiveness of HyperExpress, we conducted ablation experiments on the three modules of HyperExpress. As shown in \cref{tab:ablation1}, each module improves the model performance to a certain extent. Additionally, we present the ablation results of the margin parameter $\gamma$ and the weights $\lambda$ of each loss term in Appendix 3.
\begin{table}
  \centering
  \caption{Ablation study on HyperExpress model components on the $D1$ dataset \cite{cendra2025ICE}.}
  \label{tab:ablation1}
  \resizebox{\columnwidth}{!}{
    \begin{tabular}{ccc|cccc}
      \toprule
      HCL& HEL & HP & SIM$^I$ & SIM$^C$ & ACC$^1$ & ACC$^3$ \\
     Module & Module & Module & (\%) & (\%) & (\%) & (\%) \\
      \midrule
      \cmark & \xmark & \xmark & $0.625$ & $0.769$ & $0.326$ & $0.509$ \\
      \cmark & \cmark & \xmark & $0.688$ & $0.771$ & $0.330$ & $0.518$ \\
      \cmark & \xmark & \cmark & $0.621$ & $0.765$ & $0.348$ & $0.522$ \\
      \midrule
      \rowcolor{blue!5!white} 
      \cmark & \cmark & \cmark & \textbf{0.699} & $\textbf{0.786}$ & $\textbf{0.504}$ & $\textbf{0.736}$ \\
      \bottomrule
    \end{tabular}
  } 
\end{table}

%% file: sec/6_conclusion.tex
\section{Conclusion}
This paper proposes the task of Compositionally Interpretable Intrinsic Concept Extraction (CI-ICE), which aims to extract composable object-level concepts and attribute-level concepts from a single image using Text-to-Image (T2I) models. We design the HyperExpress method, which addresses the CI-ICE problem from two core aspects. In terms of concept disentanglement, we have devised a concept learning approach that can decompose complex visual concepts into object-level concepts and attribute-level concepts while preserving the hierarchical structure and associative relationships between concepts. Subsequently, regarding the compositionality of concepts, we have designed a concept optimization method to achieve concept compositionality by constraining the concept embedding space. Experimental results demonstrate that HyperExpress is an effective solution for the CI-ICE task.

\section*{Acknowledgements}
This work was supported by Guangdong Basic and Applied Basic Research Foundation under Grant 2024A1515140010, Key Areas Research and Development Program of Guangzhou under Grant 2023B01J0029,  the Guangdong Provincial Key Laboratory of Cyber-Physical System under Grant 2020B1212060069, the Science and Technology Development Fund, Macau SAR, under Grant 0079/2025/AFJ and 0193/2023/RIA3, and the University of Macau under Grant MYRG-GRG2024-00065-FST-UMDF.